\documentclass[journal]{IEEEtran} 
\IEEEoverridecommandlockouts

\usepackage[pdftex]{graphicx}

\usepackage{amsmath}
\usepackage{dblfloatfix}

\usepackage{url}

\def\BibTeX{{\rm B\kern-.05em{\sc i\kern-.025em b}\kern-.08em
    T\kern-.1667em\lower.7ex\hbox{E}\kern-.125emX}}

\begin{document}

\title{Metadata-Aware Adaptation of a Generative Foundation Model for Conditional CMR Synthesis}

\author{
    \IEEEauthorblockN{1\textsuperscript{st} Marc Rodríguez$^{1}$}
    , \IEEEauthorblockN{2\textsuperscript{nd} Grzegorz Skorupko$^{2}$}
    , \IEEEauthorblockN{3\textsuperscript{rd} Nay Aung$^{3,4}$}
    , \IEEEauthorblockN{4\textsuperscript{th} Steffen E Petersen$^{3,4}$}
    , \IEEEauthorblockN{5\textsuperscript{th} Karim Lekadir$^{2,5}$}
    , \IEEEauthorblockN{6\textsuperscript{th} Polyxeni Gkontra$^{2}$}
    \\
    \IEEEauthorblockA{\textit{Facultat de Matemàtiques i Informàtica, Universitat de Barcelona, Spain}} \\%
    \IEEEauthorblockA{\textit{Barcelona Artificial Intelligence in Medicine Lab (BCN-AIM), Facultat de Matemàtiques i Informàtica, Universitat de Barcelona, Spain}}\\
    \IEEEauthorblockA{\textit{William Harvey Research Institute, NIHR Barts Biomedical Research Centre, Queen Mary University London, Charterhouse Square, London, UK }}\\
    \IEEEauthorblockA{\textit{Barts Heart Centre, St Bartholomew’s Hospital, Barts Health NHS Trust, West Smithfield, London, UK}}\\
    \IEEEauthorblockA{\textit{Institució Catalana de Recerca i Estudis Avançats (ICREA), Passeig Lluís Companys 23, Barcelona, Spain}}\\
    \IEEEauthorblockA{polyxeni.gkontra@ub.edu}
}

\maketitle

\begin{abstract}
    Synthetic image generation is a promising strategy to address data scarcity and the underrepresentation of clinically important phenotypes in medical imaging, yet generating images that faithfully reflect meaningful patient characteristics remains challenging. In this work, we investigate metadata-conditioned cardiac magnetic resonance (CMR) synthesis using a pretrained latent diffusion model, encoding structured clinical metadata and slice position as textual prompts to guide CMR generation. To improve metadata adherence and address the imbalance of clinical attributes, we integrate three strategies: Metadata-Free Classifier-Free Guidance (CFG), Contrastive Batching, and Inverse-Frequency Sampling. The framework was fine-tuned and evaluated on 59,058 short-axis CMR from the UK Biobank using paired image similarity, distributional fidelity, and subgroup-level analyses. The combined approach achieved a Fréchet Inception Distance (FID) of 37.47, improving by 57.04\% over the same model fine-tuned without these strategies and by 28.68\% over a previous text-conditioned CMR diffusion baseline requiring cardiac geometry as additional input, while relying solely on patient metadata. This distributional gain, driven mainly by Metadata-Free CFG, came with a modest reduction in paired similarity, suggesting that the model prioritizes population-level realism over exact image reproduction. Subgroup analyses demonstrated improved alignment across demographic and acquisition-related metadata, with disease-specific conditioning being the most challenging task. These findings demonstrate the potential of generative foundation models for clinically meaningful CMR synthesis while highlighting the need for more effective metadata-aware conditioning strategies. Our code is available at \url{https://github.com/rodriguezmarc/conditional-cmr}.
\end{abstract}

\section{Introduction}
Cardiac Magnetic Resonance (CMR) is the reference modality for non-invasive cardiac assessment, playing a pivotal role in the diagnosis, prognosis, and monitoring of cardiovascular disease \cite{Pennell2010}. Building upon the rich information provided by CMR, recent advances in machine learning have demonstrated significant potential for enhancing automated CMR analysis and supporting clinical decision-making~\cite{fu2026development, wang2024screening, shad2026generalizable}. Nonetheless, privacy restrictions, the high cost of expert annotation, and the inherent scarcity of certain patient populations and disease phenotypes result in limited and imbalanced datasets~\cite{fairness_puyol, biased_classifiers, biased_classifiers2}. Thus, as medical image analysis has moved from isolated proof-of-concept models towards clinically relevant systems, the primary bottlenecks to developing robust and clinically applicable AI models have shifted from model architecture to data availability, diversity, and quality \cite{b1_zhang2022}.

\begin{figure*}[!t]
    \centering
    \includegraphics[width=0.9\textwidth, 
    clip,
    height=0.26\textheight]{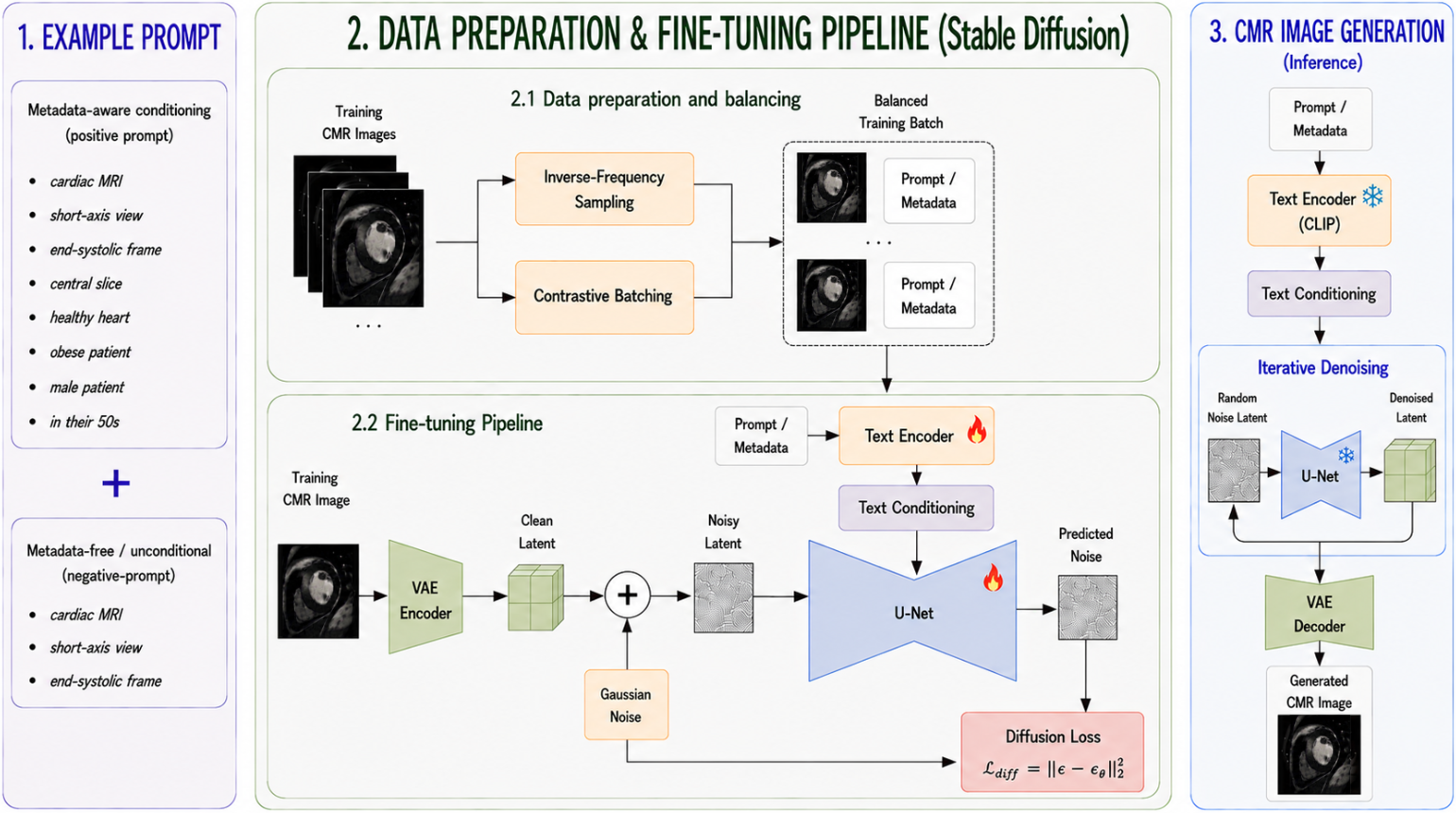}
     \vspace{-0.3cm}
    \caption{Overview of the combined conditional CMR image generation pipeline. Target CMR characteristics are encoded as text prompts and fed to a pretrained latent diffusion model, which is then fine-tuned using the proposed techniques to improve metadata adherence and mitigate class imbalance during training.}
    \label{fig:pipeline}
\end{figure*}

Synthetic image generation has emerged as a practical strategy for addressing these limitations, thus supporting data augmentation \cite{Koetzier2024-jh}, privacy-preserving experimentation \cite{DuMont_Schutte2021} and fairness-aware dataset balancing \cite{Ktena2024}. Recent diffusion models have substantially improved the realism and diversity of synthetic medical images \cite{kaleta2024minimal}. However, visual plausibility alone is insufficient; clinically meaningful synthesis also requires generated images to faithfully reflect patient metadata. Recently, the MINIM framework~\cite{b5_wang2025} demonstrated that a single text-conditioned generative model can synthesize clinically useful images across multiple modalities and anatomical sites, improving downstream clinical tasks. However, MINIM was not evaluated on CMR, and metadata-conditioned synthesis, as demonstrated for other anatomies such as the brain~\cite{Pinaya2022-mi}, remains largely unexplored.  

In this work, building on the MINIM text-conditioning paradigm, we adapt a general-domain latent diffusion model to the CMR domain by encoding structured patient metadata together with slice position as textual prompts. Unlike MINIM, which relies on large-scale multimodal medical pretraining and a self-improving reinforcement loop, we fine-tune from public Stable Diffusion weights, and focus on metadata-aware conditioning and sampling strategies to mitigate class imbalance, enabling the generation of clinically meaningful short-axis CMR images. The proposed framework improved distributional fidelity (FID) over both a fine-tuned baseline and a previous CMR diffusion baseline \cite{b2_skorupko2026} requiring cardiac geometry, while relying solely on metadata. We further analyze how conditioning reliability varies across metadata attributes with differing levels of class imbalance.

\section{Methods}
The work integrates three complementary metadata-aware strategies into a latent diffusion pipeline for conditional CMR synthesis. Metadata-Free Classifier-Free Guidance strengthens metadata conditioning during inference, Contrastive Batching increases metadata diversity within training batches, and Inverse-Frequency Sampling improves exposure to underrepresented metadata values during data loading. The overall workflow is illustrated in Fig.~\ref{fig:pipeline}.

\begin{table*}[!t]
\caption{Paired and distributional fidelity metrics together with the MAE and MS-SSIM diversity ratios (DR). Standard deviations for the DR were obtained by first-order error propagation. $\downarrow$/$\uparrow$ indicate that lower/higher values are better; for DR, values closer to 1 ($\rightarrow$1) indicate agreement with the reference distribution. Metadata-aware strategies are added incrementally on the fine-tuned Stable Diffusion (SD) baseline. Abbreviations: MF-CFG, Metadata-Free Classifier-Free Guidance; IFS, Inverse-Frequency Sampling; CB, Contrastive Batching.}
\vspace{-0.1cm}
\setlength{\tabcolsep}{6pt}
\centering
\footnotesize
\begin{tabular}{lcccccc}
\hline
\textbf{Metric} &
\textbf{Fine-tuned SD} &
\textbf{MF-CFG} & 
\textbf{MF-CFG+IFS} &
\textbf{MF-CFG+CB} & 
 \textbf{MF-CFG+IFS+CB} &
\textbf{Skorupko et al. \cite{b2_skorupko2026} }
\\
\hline
MAE $\downarrow$ & $0.257 \pm 0.040$ &  $0.254 \pm 0.045$ & $0.261 \pm 0.041$ & $0.263 \pm 0.042$  & $0.260 \pm 0.043$& $\mathbf{0.253} \pm 0.040$ \\
MS-SSIM $\uparrow$ & $0.192 \pm 0.075$ & $0.177 \pm 0.079$ & $0.164 \pm 0.072$ & $0.163 \pm 0.073$  & $0.171 \pm 0.076$ & $\mathbf{0.266} \pm 0.075$ \\
FID $\downarrow$ & $87.22$ & $44.84$ & $39.12$ & $\mathbf{37.08}$ & $ 37.47$ & $52.54$  \\
FRD$\downarrow$ & $4.76$ & $4.64$ & $\mathbf{4.27}$ & $4.44$ & $4.64$ & $5.65$ \\
MAE DR$\rightarrow$1 & $0.931 \pm 0.189 $ & $1.010 \pm 0.127$ & $\mathbf{1.006} \pm 0.178$ & $1.050 \pm 0.185$ & $1.026 \pm 0.191$ & $1.248 \pm 0.178$ \\
MS-SSIM DR$\rightarrow$1 & $0.940 \pm 0.126$ & $1.081 \pm 0.197$ & $\mathbf{1.0323} \pm 0.120$ & $1.056 \pm 0.122$ & $1.063 \pm 0.126$ & $1.068 \pm 0.111$  \\
\hline
\end{tabular}
\label{tab:setup3_global}
\end{table*}

\subsection{Metadata-Free Classifier-Free Guidance (MF-CFG)}

A common inference-time technique for Stable Diffusion-based architectures to strengthen the conditioning effect is Classifier-Free Guidance (CFG), which combines conditional and unconditional predictions so that the model is optimized according to the difference of both results ~\cite{b13_ho2022}. The proposed MF-CFG replaces the unconditional branch with a metadata-free version of the input prompt, obtained by removing metadata segments while preserving the acquisition context. The metadata-aware and a metadata-free prompt are then passed as the positive and negative prompts respectively, encouraging the guidance term to emphasize metadata-specific information.

\subsection{Contrastive Batching (CB)}
Batch training not only defines how many samples are processed at each optimizer step, but also sample diversity. In conditional generation, random sampling from an imbalanced dataset may produce batches dominated by frequent metadata values. Therefore, changing the order of sampled indices alters batch composition and increases exposure to less frequent conditioning values \cite{b21_zhao2014, b22_yang2023}. CB uses this approach to construct batches that contain different values from a selected metadata-attribute whenever sufficient diversity is available, increasing within-batch metadata variation. Metadata values from the selected attribute are sampled without replacement, with one prompt randomly drawn from the corresponding metadata bucket for each selected value. If the batch size exceeds the number of available values, all values are sampled once before being reused cyclically.

\subsection{Inverse-Frequency Sampling (IFS)}
Because a sample may be rare for one metadata attribute but common for another in multi-class conditional generation, balancing is performed at metadata level rather than the sample level, as any weighting system would be ambiguous and inexact. Inverse-Frequency Sampling computes weights from the inverse metadata value frequencies to increase the representation of underrepresented values \cite{b2_skorupko2026}. Let $c$ be a given batch-selected metadata attribute, and $x \in X_c$ be one of its possible values. Furthermore, let $N_x$ be the number of samples from the training split that contain the $x$ value for the $c$ attribute. Its inverse-frequency weight $w_x$ is:
\begin{equation}
    w_x = \min(\frac{\bar{N}_c}{N_x}, \tau)
\end{equation}
where $\bar{N}_c$ is the mean bucket size across all values of $c$. The value $w_x$ was capped at $\tau=2.0$ to prevent extremely rare values from affecting results. During batch construction, metadata values for the selected attribute are sampled proportionally to these weights, prioritizing minority values. Specifically, for a given attribute $c$, the probability of sampling sample $n_i$ with metadata value $x_i^c$  is:
\begin{equation}
P(n_i\mid c)
=
P(x_i^c \mid c)
\cdot
P(n_i \mid x_i^c)
= 
\frac{w_{x_i^c}}{\sum_{x \in X_c} w_x}
\cdot
\frac{1}{N_{x_i^c}}
\end{equation}

Since the active metadata attribute varies across batches, balancing is applied across all conditioning metadata during training~\cite{b2_skorupko2026}. Combined with CB, value selection remains contrastive while being weighted by inverse-frequency probabilities, prioritizing minority values.  

\section{Implementation}

\subsection{Dataset and Prompt Construction}

The dataset used in this work originates from the UK Biobank \cite{Sudlow2015}. For each subject, the mid-ventricular short-axis slice together with two adjacent slices were extracted from the end-systolic frames of the cine CMR acquisition, along with the corresponding metadata. Samples were then split into training and testing sets at the patient level using a $90:10$ ratio, resulting in $53,298$ training images and $5,760$ testing.

CMR generation is formulated as a text-conditioned task that uses prompts to describe image characteristics and patient metadata \cite{b5_wang2025}. Each sample is transformed into a standardized record consisting of the processed image path and a textual prompt beginning with a fixed acquisition context (imaging modality, view, and cardiac phase) and followed by patient metadata: slice position (mid-ventricular $\pm 1$), pathology (healthy, heart failure, myocardial infarction, ischemic heart disease, atrial fibrillation), BMI group (underweight: $< 18.5$, normal: $18.5 - 24.9$, overweight: $25.0-29.9$, obese: $\geq 30.0\ \mathrm{kg}/\mathrm{m}^2$), sex (male, female), and age group (40s, 50s, 60s, 70s, 80s). An example prompt is: \textit{Cardiac MRI, short-axis view, end-systolic frame, mid-ventricular + 1, healthy heart, obese patient, male patient, patient in their 50s}. 

The dataset exhibits substantial class imbalance across several metadata attributes. This is most pronounced for pathology, where healthy subjects dominate, and for BMI and age, where underweight and extreme age groups are underrepresented. In contrast, sex is relatively balanced, while slice position is inherently balanced. These imbalances motivate the metadata-aware strategies proposed in this work to improve the representation of minority metadata values.

\subsection{Model and Training Workflow}
In this work, we built on the released MINIM text-conditioning approach and codebase \cite{b5_wang2025}. As the pretrained MINIM weights are not publicly available, we initialized the model from publicly available Stable Diffusion weights (\texttt{CompVis/stable-diffusion-v1-4}, Hugging Face) \cite{Rombach_2022_CVPR} and fine-tuned it combining the proposed metadata-aware strategies. Each prompt is injected as conditioning and paired with its reference image. The implementation follows a Stable Diffusion-like latent diffusion architecture: a frozen VAE maps sample images to latent space, the U-Net is fine-tuned to denoise CMR latents, and a text encoder provides prompt embeddings to the denoising process \cite{Rombach_2022_CVPR}. During inference, synthetic images are generated for the testing split, paired with the corresponding reference images, and evaluated using quantitative metrics.

\begin{figure*}[!t]
\centering

\includegraphics[
    width=\textwidth,
    trim={0 0.7cm 0 0cm},
    clip
]{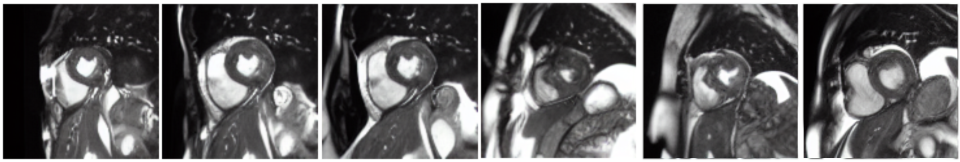}
\noindent
\makebox[\textwidth][c]{%
\begin{minipage}{0.1667\textwidth}
\centering\footnotesize\textbf{Underweight}
\end{minipage}%
\begin{minipage}{0.1667\textwidth}
\centering\footnotesize\textbf{Normal BMI}
\end{minipage}%
\begin{minipage}{0.1667\textwidth}
\centering\footnotesize\textbf{Overweight}
\end{minipage}%
\begin{minipage}{0.1667\textwidth}
\centering\footnotesize\textbf{Obese Seed 1}
\end{minipage}%
\begin{minipage}{0.1667\textwidth}
\centering\footnotesize\textbf{Obese Seed 2}
\end{minipage}%
\begin{minipage}{0.1667\textwidth}
\centering\footnotesize\textbf{Obese Seed 3}
\end{minipage}}
  \vspace{-0.6cm}
\caption{Representative generated CMR images by BMI category showing the conditioning effect on generation. From underweight to obese, the generated images qualitative show progressively greater adipose tissue, consistent with increasing BMI. The three rightmost images were generated from the same obese conditioning prompt using different random seeds, demonstrating the diversity achievable under identical conditioning.}
\label{fig:samples}
\end{figure*}

\begin{figure*}[!b]
    \includegraphics[
        width=\textwidth,
        trim=0 0.4cm 0 0.3cm,
        clip
    ]{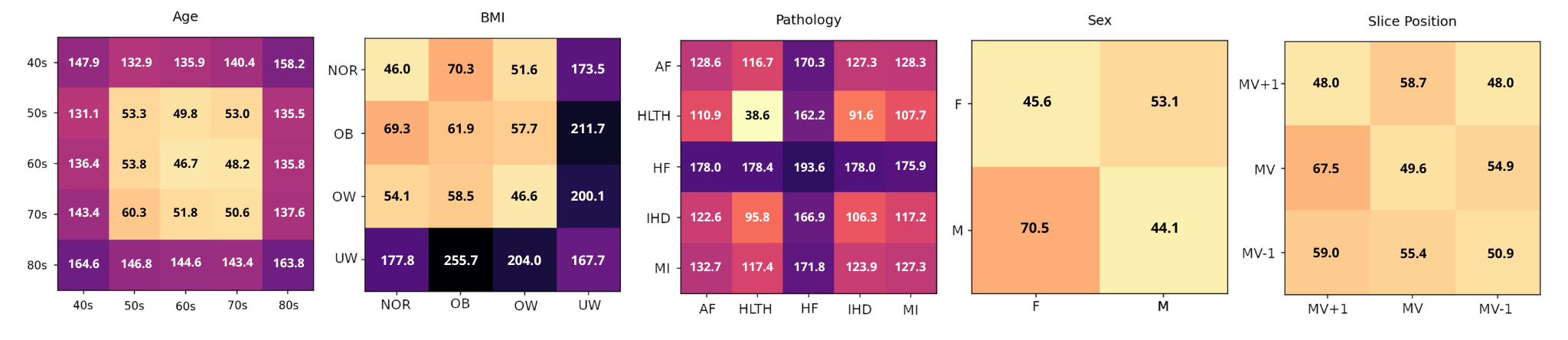}
     \vspace{-0.8cm}
    \caption{Subgroup distributional fidelity (FID) matrices for each metadata attribute. Columns correspond to generated subgroups and rows to reference subgroups. Abbreviations: NOR, normal; OB, obese; OW, overweight; UW, underweight; AF, atrial fibrillation; HLTH, healthy; HF, heart failure; IHD, ischemic heart disease; MI, myocardial infarction; F, female; M, male; MV, mid-ventricular. }
    \label{fig:subgroup_fid_matrices}
\end{figure*}

\subsection{Evaluation Protocol}
The evaluation considers three complementary aspects: paired-image fidelity, distributional fidelity, and subgroup performance. Paired fidelity compares each generated image with its reference image using Mean Absolute Error (MAE), which quantifies pixel-level differences, and Multi-Scale Structural Similarity (MS-SSIM)~\cite{wang2003multiscale}, used to investigate anatomical similarities. Distributional fidelity compares generated and real testing sets using the Fréchet Inception Distance (FID)~\cite{fid} and the Fréchet Radiomics Distance (FRD)~\cite{Konz_2026}, while subgroup analyses evaluate model performance across the different metadata categories. In addition, distributional matrices compare generated and real image distributions across metadata values. The diagonal entries compare generated and real samples with the same metadata value, while off-diagonal values reveal potential overlap or poor separation between metadata categories. Finally, diversity ratios are computed for the pair fidelity metrics, indicating variation in comparison to the reference testing split.

\section{Results and Discussion}

\subsection{Overall Performance}
Table~\ref{tab:setup3_global} summarizes the global results. The model incorporating the combined metadata-aware strategies achieved an FID of $37.47$, improving by $57.04\%$ over the same latent diffusion model fine-tuned without the proposed strategies and by $28.68\%$ compared with the text-conditioned CMR diffusion model of Skorupko et al. \cite{b2_skorupko2026}, which requires cardiac geometry as additional conditioning input. Ablation results indicate that MF-CFG drives most of the distributional improvement (FID: 87.22 to 44.84), with the sampling strategies providing further gains. MF-CFG + CB achieved the lowest FID (37.08), while MFCFG + IFS achieved the lowest FRD (4.27). As the margin between the combined
model and the best single-strategy variants is small, we adopt the combined configuration as the representative model for the subgroup and diversity analyses that follow, since it balances distributional and paired fidelity rather than optimizing either metric in isolation.

In the combined model, paired fidelity metrics, with MAE equal to $0.260$ and MS-SSIM equal to $0.171$, indicate that it preserves the short-axis CMR appearance without behaving as an exact reconstruction model. As it conditions on metadata alone, it is not intended to reproduce a specific reference image; the higher paired similarity of Skorupko et al. \cite{b2_skorupko2026}, which additionally uses cardiac geometry, reflects that extra spatial information. Distributional fidelity (FID, FRD) is therefore better suited to this metadata-conditioned setting. Moreover, the near-unity results of both diversity ratios indicate that the generated set preserves realistic variation without collapsing into a narrow set of similar images. Representative generated samples for different BMI classes are shown in Fig.~\ref{fig:samples}.

\subsection{Subgroup Performance Analysis}
The distributional FID matrices in Fig.~\ref{fig:subgroup_fid_matrices} summarizes subgroup-level performance across metadata attributes for the combined model, with lower diagonal values indicating better agreement between the generated and real distributions. Diagonal mean subgroup FID ranged from $44.90$ for sex and $49.48$ for slice position to $80.55$ for BMI and $92.45$ for age, while pathology exhibited the highest value ($118.87$), indicating that even same-subgroup generation is least accurate for disease categories. For every attribute, the diagonal mean FID was lower than the corresponding off-diagonal value ($61.83$ for sex, $57.25$ for slice position, $129.53$ for BMI, $115.17$ for age, and $138.68$ for pathology), confirming that generated subgroups align more closely with their matching real subgroups than with different ones. These results indicate that conditioning is most reliable for well-balanced metadata attribute. However, pathology is also the most underrepresented attribute, so part of the elevated FID may reflect the known sensitivity of FID to small sizes~\cite{chong} rather than conditioning difficulty alone. Overall, subgroup performance closely followed the underlying data distribution, with the poorest results consistently corresponding to the most underrepresented metadata values, indicating that the applied strategies substantially improve generation but do not fully overcome class imbalance.

\section{Conclusion}
Building on the MINIM text-conditioning paradigm and public Stable Diffusion weights, we proposed a metadata-aware framework for conditional CMR synthesis that conditions image generation on structured patient metadata and slice position without requiring cardiac geometry as input. The framework improved distributional fidelity over both a fine-tuned Stable Diffusion baseline  and a previously published CMR text-conditioned diffusion baseline while generating realistic and diverse short-axis CMR images. Subgroup analyses further showed that conditioning is more reliable for balanced metadata attributes, such as sex and slice position, than for highly imbalanced clinical variables, particularly pathology. The ablation results showed that MF-CFG was the main driver of distributional improvement, while the balancing strategies provided complementary refinements. These results demonstrate the potential of latent diffusion models for metadata-aware CMR synthesis while highlighting that, although the proposed strategies improve distributional fidelity, robust and fully controllable metadata conditioning under class imbalance remains an open challenge. Future work will assess the effectiveness of the generated images in downstream CMR analysis tasks, particularly for improving the performance and fairness of models trained on underrepresented patient subgroups.

\section*{Acknowledgment}

This work is part of the project TrustAI-ES (PID2023-146751OA-I00), funded by MICIU/AEI/10.13039/501100011033.  This work received funding from the European Union’s Horizon Research and Innovation program under Grant Agreement No. 101080430 (AI4HF project). This work is also supported by the European Union’s Horizon Europe research and innovation program under Grant Agreement No. 101057849 (DataTools4Heart project). This work was conducted using the UK Biobank resource under access application 2964.

\ifCLASSOPTIONcaptionsoff
  \newpage
\fi


%
%
%
\newpage
\bibliographystyle{IEEEtran}
\bibliography{references}
\end{document}